# A Cost-Effective Test Bench for Evaluating Safe Human-Robot Interaction in Mobile Robotics

Atefeh Fereydooni
*Faculty of Electrical Engineering*
*K. N. Toosi University of Technology*
Tehran, Iran
atefefereydouni@email.kntu.ac.ir

Armin Azarakhsh
*Faculty of Electrical Engineering*
*K. N. Toosi University of Technology*
Tehran, Iran
armin.azarakhsh@email.kntu.ac.ir

Ayda Shafiei
*Faculty of Electrical Engineering*
*K. N. Toosi University of Technology)*
Tehran, Iran
ayda.shafiei@email.kntu.ac.ir

Hesam Zandi
*Department of Electronics*
*Faculty of Electrical Engineering*
*K. N. Toosi University of Technology*
Tehran, Iran
zandi@kntu.ac.ir

Mehdi Delrobaei
*Department of Mechatronics*
*Faculty of Electrical Engineering*
*K. N. Toosi University of Technology*
Tehran, Iran
delrobaei@kntu.ac.ir

*Abstract*—Safety concerns have risen as robots become more integrated into our daily lives and continue to interact closely with humans. One of the most crucial safety priorities is preventing collisions between robots and people walking nearby. Despite developing various algorithms to address this issue, evaluating their effectiveness on a cost-effective test bench remains a significant challenge. In this work, we propose a solution by introducing a simple yet functional platform that enables researchers and developers to assess how humans interact with mobile robots. This platform is designed to provide a quick yet accurate evaluation of the performance of safe interaction algorithms and make informed decisions for future development. The platform's features and structure are detailed, along with the initial testing results using two preliminary algorithms. The results obtained from the evaluation were consistent with theoretical calculations, demonstrating its effectiveness in assessing human-robot interaction. Our solution provides a preliminary yet reliable approach to ensure the safety of both robots and humans in their daily interactions.

## I. Introduction

Over the past few years, collaborative robots have steadily made their way into human settings. Their benefits were especially evident during the COVID-19 pandemic, as they facilitated social distancing and reduced the need for physical interaction. As time passes, there is a growing acknowledgment of the feasibility of collaborative robots (cobots), resulting in increased integration across various industries [1].

Hence, it is becoming a new norm for humans and robots to share the same space and work closely together. Some studies have investigated the relationship between the use of industrial robots and workplace harm and have concluded that applying industrial robots is closely associated with harm reduction [1]. The statistics of human injuries caused by interacting with robots have been reviewed in various articles [2]–[5]. Some cases have estimated the damage caused by human-robot collisions using different criteria, and others have focused on the dangers of the imminent arrival of robots that interact with pedestrians on the ground and mix with the human population. These robots consist of semi-autonomous or fully autonomous mobile platforms designed to provide several services, such as assistance, patrol, tour guide, delivery, and human transportation [6].

A study conducted over 26 years in the United States found 41 robot-related deaths. Most cases involved stationary robots (83%) and robots operating autonomously (78%) striking the deceased. A significant part of these incidents occurred during the repair and maintenance of the robot [7].

As the presence of robots in human environments continues to grow, the development of secure interaction algorithms is becoming increasingly crucial in robotics. Ensuring the safety of both humans and robots is of utmost importance, and proper interaction algorithms can aid in achieving this goal. Thus, researchers and developers must create algorithms prioritizing safety in human-robot interactions. Doing so can foster a future where robots play a constructive societal role. Many researchers have designed and presented different mobile platforms and plenty of algorithms. Still, the need for a test bench with a simple and cost-effective structure to validate relevant algorithms remains unmet [8]–[12].

Our primary goal was to devise an economical and straightforward testing configuration and present preliminary results on collision avoidance. The proposed platform consists of two parts, an upper body, and a lower body, the lower body of which is a mobile base and the upper body of which is a mechanical arm with two degrees of freedom.

The platform also includes proximity sensors that detect the distance to the obstacles and a processor that analyzes the data and sends commands to the upper or lower body. Two preliminary algorithms that prioritized the movement of the base or the arm were finally implemented.

This paper is organized as follows. Section II describes the architecture and the methodology employed in this study. The outcomes of the suggested test bench are outlined in Section III, while the concluding remarks are provided in Section IV.

## II. METHOD

In this section, we first present the general structure of the proposed platform to form the suggested test bench. The platform consists of a processor unit, a communication unit, proximity sensors, a mobile base, and a 2 DoF mechanical arm. Then, we explain the structure of each of these components. The implemented algorithms are introduced, and the predicted and obtained results are compared. The diagram in Fig. 1 depicts the concept of the testing platform.

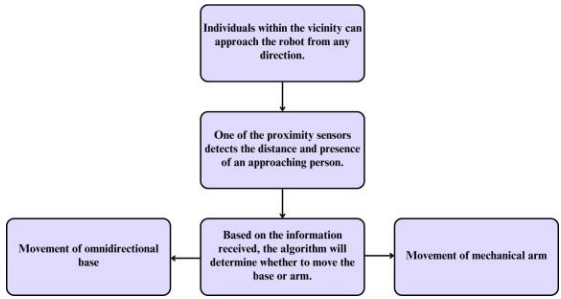

Fig. 1: The conceptual diagram of the test bench.

### A. General Architecture

The proposed platform comprises four primary components: the processor unit, communication link, distance sensor pack, and body. The sensor pack measures distances and facilitates communication between the platform and its surroundings. The upper body of the platform, which features a mechanical arm, can move in two directions and is powered by two servo motors. The base of the platform is omnidirectional, allowing it to move in any direction within its environment. Further details on these components can be found in the following section.

### B. Implementation

*1) Mobile Base:* To build the mobile base, we employed three 7.2 V DC motors (GHM-03, Lynxmotion, VT, USA) and three omnidirectional wheels coupled to each motor. The motors were attached to a circular piece of ABS plate with a 50 cm diameter. Each motors is placed at an angle of 120 degrees from each other. The use of omnidirectional wheels increases the mobility and maneuverability of the robot. This omnidirectional base has 3 degrees of freedom in x, y and $\theta$ directions.

Two L293D ICs drive DC motors; each motor can move in two clockwise or anti-clockwise modes. This mobile base can move from one point to another by turning on two motors in two opposite directions and turning off the other motor in a straight path in 6 hypothetical directions +L, -L, M, -M, +N, -N. while the third wheel, which its engine is off, slides along the direction of movement. Table I shows how each of these moves is done.

In Table I, number 1 indicates that the motor is on and rotates in the clockwise direction, and number -1 indicates that the motor is on and turns in the counterclockwise direction (Fig. 2). In the following, the motion equations of this base will be examined.

TABLE I: Direction of the base movement and wheel rotation.

| $V_{W_1}$ | $V_{W_2}$ | $V_{W_3}$ | Movement[a] |
|---|---|---|---|
| 0 | 1 | -1 | +N |
| 0 | -1 | 1 | -N |
| 1 | -1 | 0 | -M |
| -1 | 1 | 0 | +M |
| 1 | 0 | -1 | +L |
| -1 | 0 | 1 | -L |

[a]Direction of movement in hypothetical coordinates.

Fig. 3 shows the schematic view of the three-wheel mobile robot. "R" is the length from the center of the base to each wheel. $V_{W_1}$, $V_{W_2}$, $V_{W_3}$ are the linear velocities of the wheels. "r" is the radius of each wheel. From the kinematics equation all forces are divided into two components vertical 'Y' axis 'sin' component and horizontal 'X' axis 'cos' component.

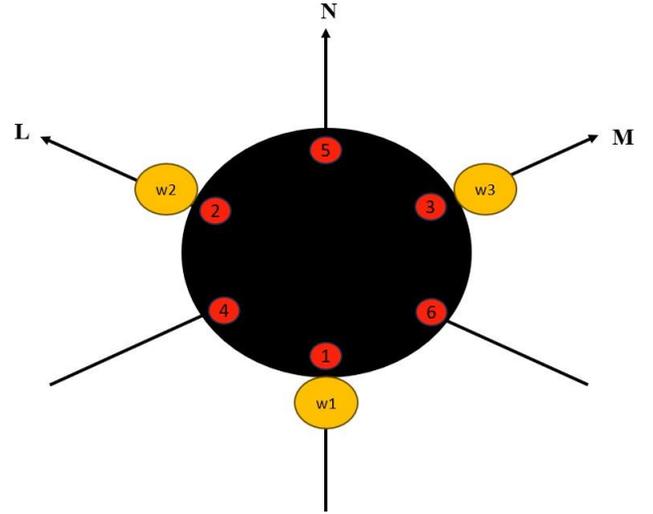

Fig. 2: Position of the sensors and wheels and the hypothetical coordinates of the base.

According to the vector analysis of the velocity of each wheel which is shown in Fig. 3, the mathematical model of omnidirectional base based on the velocity of movement of each wheel and their position is shown in (1) [13], [14].

$$\begin{bmatrix} V_{W_x} \\ V_{W_y} \\ V_\theta \end{bmatrix} = \frac{1}{3} \begin{bmatrix} -\frac{1}{2} & -\frac{1}{2} & 1 \\ \frac{\sqrt{3}}{2} & -\frac{\sqrt{3}}{2} & 0 \\ \frac{1}{R} & \frac{1}{R} & \frac{1}{R} \end{bmatrix} \begin{bmatrix} V_{W_1} \\ V_{W_2} \\ V_{W_3} \end{bmatrix} \quad (1)$$

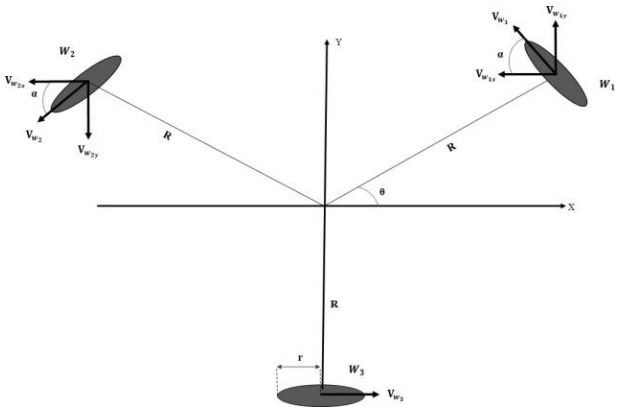

Fig. 3: schematic of a three-wheeled omnidirectional base.

*2) Directional Mechanical Arm Structure:* The arm of the platform has a simple design and can move in two directions. To achieve this, two Mg995 servo motors are used, which can rotate from 0 to 180 degrees. The first motor allows the arm to move left and right, while the second motor mounted on top of it moves the arm forward and backward. Both motors weigh 55 g and work with a voltage of 4.8-7.2 V. They require a PWM pulse with a period of 20 ms (50 Hz) to operate.

A mechanical arm is mounted on the set of these motors using a bracket installed on the second motor and three aluminum spacers, each with a height of 5 cm. The final height of this mechanical arm is 21 cm from the second servo motor. It's important to note that the arm can only bend at an angle that doesn't exceed the base's 30 cm diameter. Therefore, the movement angle of each servo is calculated accordingly. Changing the height of the arm will also change its reaction angle.

Table II shows the movement angle of each servo motor when each sensor is stimulated. The direction of arm movement, when each sensor is enabled, depends on the position of sensor number 1 in Fig. 2.

TABLE II: The angle and direction of the movement of the arm.

| Stimulated Sensor | Servo Motor 1 | Servo Motor 2 | Direction of arm movement |
|---|---|---|---|
| Sensor 1 | 90 | 45 | Front |
| Sensor 2 | 90 | 135 | Right |
| Sensor 3 | 45 | 90 | Left |
| Sensor 4 | 135 | 90 | Right |
| Sensor 5 | 90 | 135 | Behind |
| Sensor 6 | 90 | 45 | Left |

*3) Sensors Packet:* Six ultrasonic sensors, model HC-SR04, are used in this platform. The sensors are placed on the edge of the base at a distance of 60° from each other. The weight of each sensor is 10 g. This sensor has a measuring angle of 30° and a ranging distance of 2400 cm with a resolution of 0.3 cm. It also requires a working voltage of 5 V DC and a trigger pulse with a width of 10 us.

*4) Processor Unit:* The Arduino® Mega 2560 is a microcontroller board based on the ATmega2560 chip. It has 54 digital input/output pins, of which 15 can be used as PWM outputs, and 16 analog inputs. The operating voltage is 5 V, and the input voltage ranges from 7-12 V (recommended) and 6-20 V (limit). The dimensions of the board are 101.52 $mm \times$ 53.3 $mm$, and it weighs 37 g.

The proposed platform and its various components can be seen in Fig. 4.

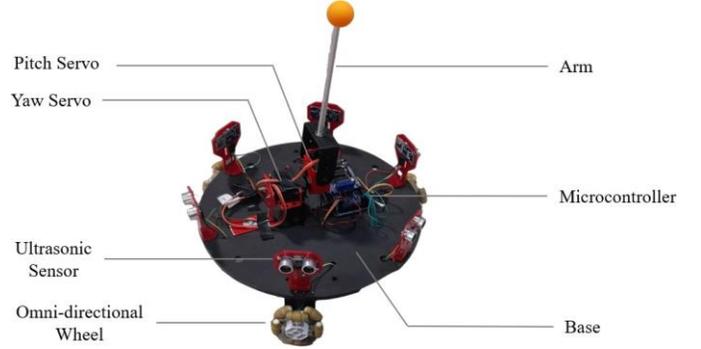

Fig. 4: The main components of the platform.

In Fig. 5, the connection of two servo motors to achieve an active universal joint with two DoF as the arm of the platform is shown. This arm can be replaced by a manipulator in more advanced robots with various capabilities.

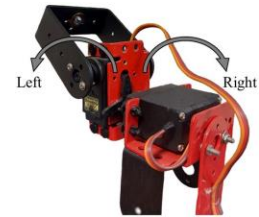

(a)

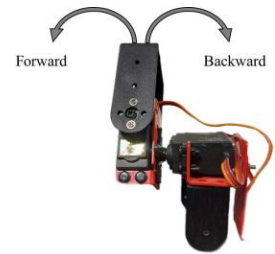

(b)

Fig. 5: The mechanical construction of the active universal joint, consisting of two servo motors. It mimics a 2 DoF inverted pendulum and acts as the manipulator, along with the base, to form a simple mobile manipulator; (a) the yaw maneuver; (b) the pitch maneuver.

*C. Proposed safe interaction algorithms*

*1) Algorithm-1:* In algorithm 1, , the given priority is to the movement of the upper body of the platform; it means that, after detecting a human or an object at an unsafe distance from itself, the platform first moves the mechanical arm in the opposite direction of the presence of that human or object. Then, by re-checking the information received from the sensors, if it still detects a person or an object in its unsafe area, it moves its base to avoid hitting and harming that person and starts moving away from them. If these conditions persist, the mobile base will continue to move in the opposite direction of the human, and the mechanical arm will remain at the same reaction angle.

---

**Algorithm 1** Algorithm for Scenario 1

1: **for** $i = 1$ to $N$ **do**          N: Number of Sensors
2:     Arm in the default position
3:     $distance_i \leftarrow$ Measured distance by sensor $i$
4:     **if** ($distance_i$ < Safe Area) **then**
5:         Arm movement(Angle, Direction)
6:         $distance_i \leftarrow$ Measured distance by sensor $i$
7:         **if** ($distance_i$ < Safe Area) **then**
8:             Base movement(Direction)
9:         **end if**
10:    **end if**
11: **end for**

---

*2) Algorithm-2:* In the algorithm2, after detecting a human being at an unsafe distance from it by the sensors of the built-in distance meter, the platform first moves its base to avoid hitting that human being and harming them and moves away from them. Then, by re-checking the information received from the sensors, if it still recognizes the human in its unsafe area, it moves the mechanical arm in the opposite direction of the presence of the human. If this condition continues, the mobile base will move in the opposite direction. The mechanical arm remains at the same reaction angle.

---

**Algorithm 2** Algorithm for Scenario 2

1: **for** $i = 1$ to $N$ **do**          N: Number of Sensors
2:     Arm in the default position
3:     $distance_i \leftarrow$ Measured distance by sensor $i$
4:     **if** ($distance_i$ < Safe Area) **then**
5:         Base movement(Direction)
6:         $distance_i \leftarrow$ Measured distance by sensor $i$
7:         **if** ($distance_i$ < Safe Area) **then**
8:             Arm movement(Angle, Direction)
9:         **end if**
10:    **end if**
11: **end for**

---

## III. EVALUATION AND EXPERIMENTAL RESULTS

*A. Calculating the velocity of the base*

In the previous section, the kinematic analysis of the omnidirectional base movement was presented, and (1) was obtained for the speed and direction of the omnidirectional base movement based on the speed and direction of rotation of the wheels. The obtained equation is used in ideal conditions. Since it is challenging to calculate the equations of motion despite the real conditions and considering the friction and the weight of the base, in this article, the speed of the base is experimentally calculated and done in such a way that, according to the limitation of the engine, the wheels spin as fast as they can.

Several time intervals for the duration of the movement of the wheels are applied to the base through the processor and software, and the amount of movement of the base has been measured for these intervals and the average speed of the base in each test has been calculated. By repeating this work in several steps and averaging the obtained $\Delta V$ s, the speed of the base movement has been calculated and used in real conditions. Experimental data can be seen in Table III.

TABLE III: The experimental data for calculating the velocity of the base in real condition.

| Test Number | $\Delta$t (ms) | $\Delta$d (cm) | $\Delta$V (cm/ms) |
|---|---|---|---|
| 1 | 2000 | 38 | 0.0190 |
| 2 | 2000 | 38.5 | 0.0192 |
| 3 | 2000 | 40 | 0.020 |
| 4 | 2500 | 50.5 | 0.020 |
| 5 | 2500 | 50 | 0.020 |
| 6 | 2500 | 50.5 | 0.020 |
| 7 | 3000 | 59.5 | 0.0198 |
| 8 | 3000 | 59 | 0.0196 |
| 9 | 3000 | 59.5 | 0.0198 |
| 10 | 5500 | 110.5 | 0.020 |
| 11 | 5500 | 111 | 0.020 |
| 12 | 5500 | 111 | 0.020 |

According to the obtained experimental data, the average aging relationship in the form of $\overline{V} = \dfrac{\sum\limits_{i=1}^{12} \Delta V_i}{12} \cong 0.02$.

*B. Evaluating the performance of this platform*

The results of the performance of this platform in each of the scenarios are figures proposed in this section. In Fig. 7 and Fig. 8, the plot shows the movement path of the human and the platform in two-dimensional coordinates x and y, and the numbers inserted in this plot indicate the time interval of each step of the human and platform reaction. In plot b, the distance between the human and the platform is displayed in terms of time, and plot c shows the reaction or non-reaction of the arm during the execution of each scenario. These plots are the results obtained from two practical tests in the laboratory environment; the conditions and how to perform them can be seen in Fig. 6.

*1) Scenario 1:* In scenario one, the primary focus is on the movement of the arm. As illustrated in Fig. 7a, the initial position of the platform and the person is at step one. The person then approaches the platform steadily, staying at a distance less than the safe distance of 50 cm. The platform detects the person being too close and, following the priority set in this scenario, moves its arm in the opposite direction of the person's movement.

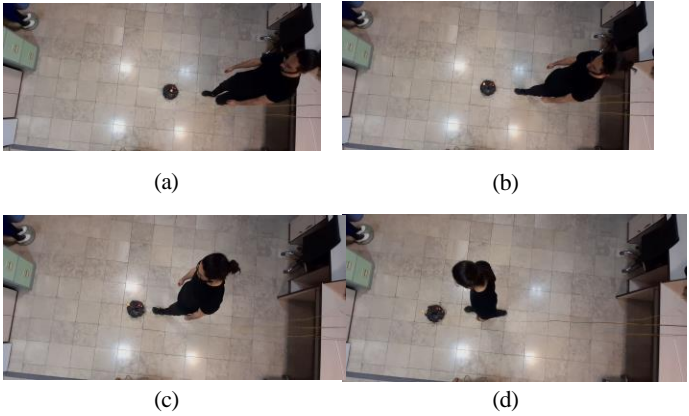

Fig. 6: Experimental tests for evaluating the performance of the platform.

This reaction happens in step 2 after that, with the continued presence of the human in the unsafe area, in addition to the arm being in a reactive angle, the base also starts to move against the direction of the human and moves away. This process continues until step 7. As shown in Fig. 7b, in steps 8 to 10, the distance between the human and the platform is more than 50 cm and the human is outside the safe range of the platform, so the base is motionless in these three steps, and the arm is in its default position.

In steps 11 to 19, the distance between the human and the platform is reduced, and the human again enters the unsafe range of the platform. Therefore, the arm is placed at a reaction angle in step 11, and 12 to 19; besides the arm must remain bent, the base must also move away from the person. But as can be seen in Fig. 7a and Fig. 7c, in steps 15 and 16, the sensors could not detect humans in the unsafe area and this platform did not detect the danger.

*2) Scenario 2:* In the second scenario, the omnidirectional base part defines the movement priority. As can be seen in Fig. 8a, the initial position of this platform and the human is at step 1, and after that the human approaches the platform at a constant speed and is at a distance less than the safe distance defined as 50 cm. The platform also detects the human at an unsafe distance; according to the priority set in the scenario, it moves its base against the direction of the human presence and moves away from him.

This process continues until step 7, and from step 7 to 9, human movement is done faster. Therefore, in step 9, the platform's arm first moves against the direction of human presence. Then the base starts to move again, and until step 11, the arm remains at the fetching angle and the base continues to move in the opposite direction.

In Fig. 8b, the distance between the human and the platform is displayed, which shows that in step 9, the distance between the human and the platform has decreased, leading to the arm's movement.

Finally, in Fig. 8c, the time that the arm has gone to the reactive angle and its duration can be seen in this plot 0 means that the arm is in the default position, and 1 means that the arm is bent in order to get away from danger.

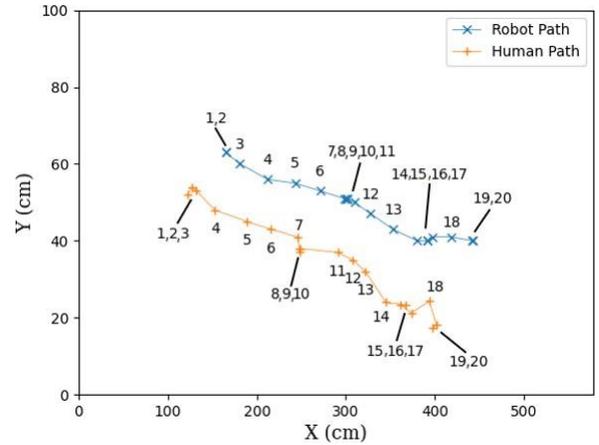

(a)

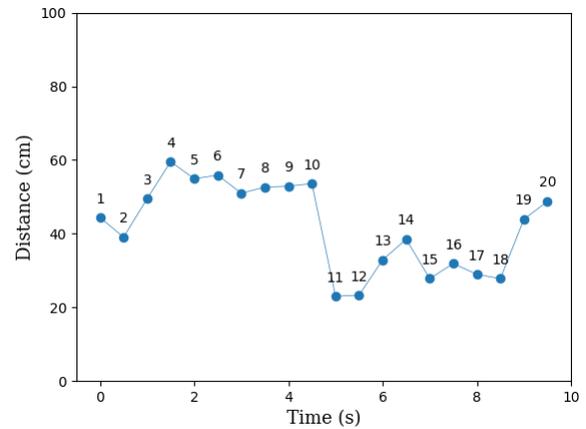

(b)

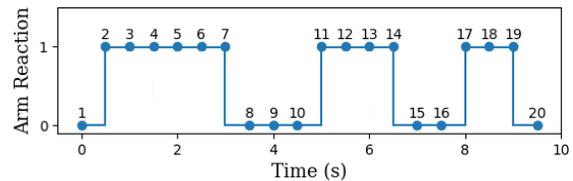

(c)

Fig. 7: The results of the performance of the proposed platform in scenario 1; (a) the path of movement of the omnidirectional base and the human; (b) the distance between the base and the human; (c) the reaction of the arm.

## IV. CONCLUSION

In this study, our goal was to design and build a cost-effective platform with simple components to implement safe human-robot interaction algorithms. After designing and building this platform, two simple algorithms of safe human-robot interaction with the aim of avoiding collisions were also designed and the performance of this platform was evaluated. The results of the evaluation of this platform were close to theoretical calculations. The proposed platform's affordable and simple design may restrict the ability to implement and evaluate more comprehensive scenarios. We have plans to improve the design in the future and our aim is to enhance its processing and interaction capabilities while still maintaining its simplicity.

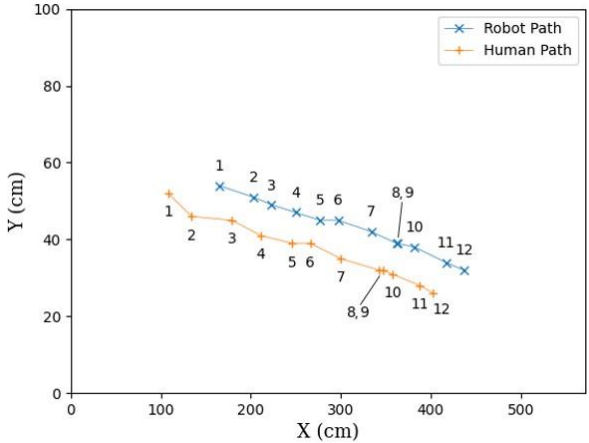

(a)

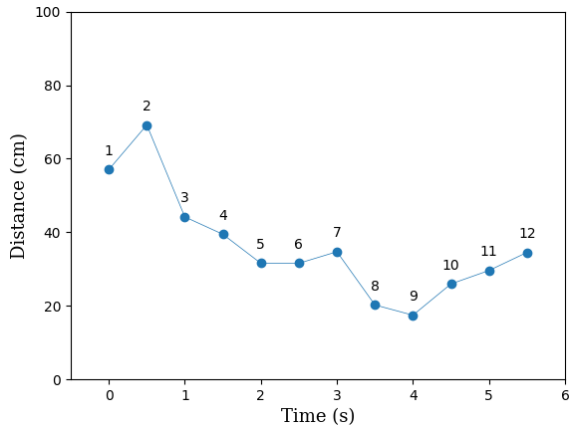

(b)

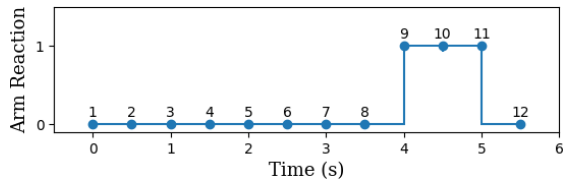

(c)

Fig. 8: The results of the performance of the proposed platform in scenario 2; (a) the path of movement of the omnidirectional base and the human; (b) the distance between the base and the human; (c) the reaction of the arm.